\documentclass[letterpaper]{article} 
\usepackage{fixltx2e} 
\usepackage{xcolor} 
\usepackage{amsmath}
\usepackage{amssymb}
\usepackage{booktabs}
\usepackage{multirow}
\usepackage{tikz}
\usepackage{aaai2026}  
\usepackage{times}  
\usepackage{helvet}  
\usepackage{courier}  
\usepackage[hyphens]{url}  
\usepackage{graphicx} 
\usepackage{natbib}  
\usepackage{caption} 
\usepackage{algorithm}
\usepackage{algorithmic}
\usepackage{amsmath}
\usepackage[table]{xcolor}
\usepackage{booktabs}
\usepackage{multirow}
\usepackage{caption}
\usepackage{enumitem}

\newcommand*\circled[1]{\tikz[baseline=(char.base)]
  \node[shape=circle,draw,line width=1pt,inner sep=0.8pt] (char) {#1};}

\newcommand*\circlednum[1]{\tikz[baseline=(char.base)]
  \node[shape=circle,draw,line width=1pt,inner sep=0.8pt] (char) {\textbf{#1}};}

\newcounter{corfn}
\newcommand{\corresponding}{%
  \ifnum\value{corfn}=0
    \footnote{Corresponding authors}%
    \setcounter{corfn}{\value{footnote}}%
  \else
    \footnotemark[\value{corfn}]%
  \fi
}

\usepackage{newfloat}
\usepackage{listings}
\DeclareCaptionStyle{ruled}{labelfont=normalfont,labelsep=colon,strut=off} 
\floatstyle{ruled}
\newfloat{listing}{tb}{lst}{}
\floatname{listing}{Listing}
\title{Causal Tracing of Object Representations in Large Vision Language Models: Mechanistic Interpretability and Hallucination Mitigation}
\author{
    Qiming Li\textsuperscript{\rm1}\equalcontrib, Zekai Ye\textsuperscript{\rm1}\equalcontrib, Xiaocheng Feng\textsuperscript{\rm1,\rm2}\corresponding, Weihong Zhong\textsuperscript{\rm1}, Weitao Ma\textsuperscript{\rm1}, Xiachong Feng\textsuperscript{\rm3}\corresponding
}
\affiliations{
    \textsuperscript{\rm 1}Harbin Institute of Technology,
    \textsuperscript{\rm 2}Peng Cheng Laboratory,
    \textsuperscript{\rm 3}The University of Hong Kong \\
    \{qmli,zkye\}@ir.hit.edu.cn

}

\usepackage{bibentry}

\begin{document}

\maketitle

\begin{abstract}
Despite the remarkable advancements of Large Vision-Language Models (LVLMs), the mechanistic interpretability remains underexplored. 
Existing analyses are insufficiently comprehensive and lack examination covering visual and textual tokens, model components, and the full range of layers.
This limitation restricts actionable insights to improve the faithfulness of model output and the development of downstream tasks, such as hallucination mitigation. 
To address this limitation, we introduce \textbf{Fine-grained Cross-modal Causal Tracing (FCCT)} framework, which systematically quantifies the causal effects on visual object perception.
FCCT conducts fine-grained analysis covering the full range of visual and textual tokens, three core model components including multi-head self-attention (MHSA), feed-forward networks (FFNs), and hidden states, across all decoder layers.
Our analysis is the first to demonstrate that MHSAs of the last token in middle layers play a critical role in aggregating cross-modal information, while FFNs exhibit a three-stage hierarchical progression for the storage and transfer of visual object representations. Building on these insights, we propose \textbf{Intermediate Representation Injection} \textbf{(IRI)}, a training-free inference-time technique that reinforces visual object information flow by precisely intervening on cross-modal representations at specific components and layers, thereby enhancing perception and mitigating hallucination.
Consistent improvements across five widely used benchmarks and LVLMs demonstrate \textbf{IRI} achieves state-of-the-art performance, while preserving inference speed and other foundational performance. 
\end{abstract}


\section{Introduction}

Large Vision-Language Models (LVLMs) have rapidly evolved, demonstrating impressive capabilities across diverse tasks. 
However, existing interpretability studies fall short in capturing the full complexity of visual information flow, thereby limiting progress in critical downstream applications such as hallucination mitigation.
Fundamental questions require further investigation, particularly regarding how LVLMs process visual object features and align them with textual semantics in cross-modal representations.
Furthermore, how visual and textual tokens elicit distinct functional behaviors from three core model components—such as multi-head self-attention (MHSA), feed-forward networks (FFNs), and hidden states—across layers, remains underexplored.
Previous studies have partially addressed these questions but still leave notable limitations. Attention knockout experiments \cite{neo2024towards} demonstrate that LVLMs extract object information from visual object tokens in the middle to late layers. However, it lacks an analysis of the cross-modal interactions between visual and textual tokens, as well as the functional roles of the MLP and hidden states. NOTICE \cite{golovanevsky2024vlms} introduces semantic image pairs for image corruption and symmetric token replacement for text corruption to analyze how MHSA and MLP contribute to information aggregation in textual tokens, but overlooks the effect of visual tokens. 

\begin{figure}[!t]
\centering
\includegraphics[width=0.45\textwidth]{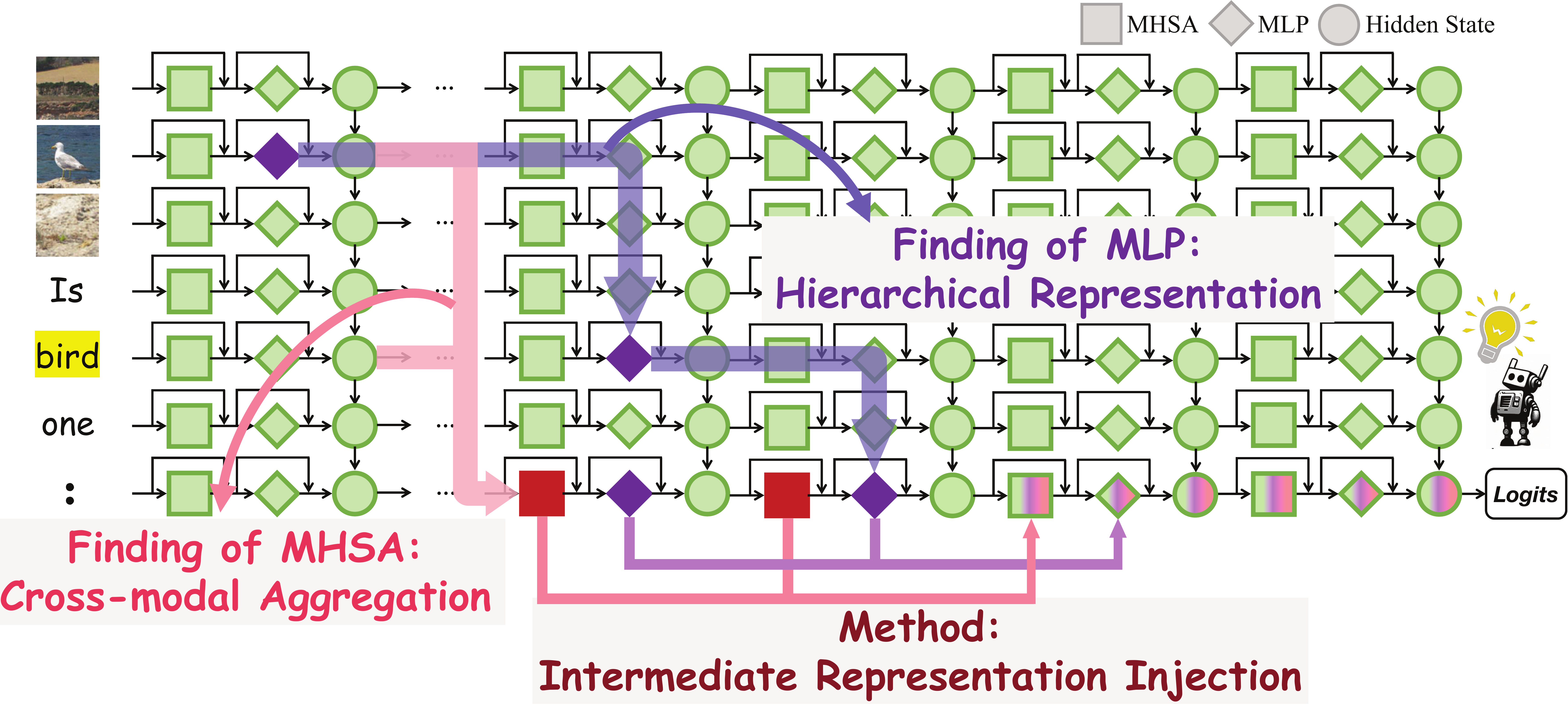} 
\caption{An overview of our proposed \textbf{Fine-grained Cross-modal Causal Tracing (FCCT)} findings and \textbf{Intermediate Representation Injection (IRI)} method. }
\label{intro}
\end{figure}

To address these limitations, we propose a \textbf{Fine-grained Cross-modal Causal Tracing (FCCT)} framework, which systematically analyzes cross-modal causal effects on visual perception by examining visual and textual tokens categorized with their position and semantic role in the input sequence, covering three core model components across layers. 
By introducing controlled Gaussian perturbations to input images, we induce measurable drops in output probabilities for existing objects. 
Then we restore specific activations using the clean activations from the original image input.
By quantifying the recovery in the LVLM's output probabilities, we precisely estimate the causal effect on visual object perception for each core components across token types and layers.
\textbf{FCCT} is the first to demonstrate that the MHSAs of the last token in middle layers play a critical role in aggregating crucial object visual and textual information, as well as the FFNs exhibit a three-stage hierarchical progression for the storage and transfer of visual object representations. 

\textbf{FCCT} not only reveals cross-modal information flow of visual objects, but also provides valuable guidance for hallucination mitigation. 
Prior studies \cite{tang2025mitigating} suggest that deep unidirectional information flow can lead to the progressive degradation of fine-grained semantic cues encoded in earlier layers, which may be a potential cause of object hallucination.
To prevent mid-layer degradation of critical information during forward and reinforce components with strong causal effects identified by \textbf{FCCT}, we further propose \textbf{Intermediate Representation Injection (IRI)}, a training-free inference-time technique that injects crucial mid-layer representations into subsequent layers, thereby enhancing visual perception capability and mitigating hallucination.
\textbf{FCCT} offers fine-grained and quantitative guidance for selecting model components and layers, serving as a theoretical foundation for the design and implementation of the \textbf{IRI} method.
Consistent improvement across five widely used benchmarks and five advanced LVLMs demonstrates that \textbf{IRI} achieves state-of-the-art (SOTA) performance.

In summary, our main contributions are three-fold:
\begin{itemize}
    \item We propose \textbf{FCCT}, a fine-grained causal analysis that covers all types of visual and textual tokens, three core components across the full layer range, providing a comprehensive mechanistic interpretability study of LVLMs.
    \item We propose \textbf{IRI}, a training-free inference-time method that effectively mitigates hallucination while preserving inference speed and other foundational capabilities.
    \item Consistent improvements across five widely used benchmarks and LVLMs not only demonstrate \textbf{IRI}'s SOTA performance, but also validate the findings of \textbf{FCCT}.
\end{itemize}

\section{Related Work}

\subsubsection{Mechanistic Interpretability of LVLMs}
While LVLMs have demonstrated remarkable capabilities across various downstream tasks, their mechanistic interpretability remains underexplored. Existing interpretability methods, such as probing \cite{salin2022vision}, activation patching \cite{basu2024understanding,palit2023towards,golovanevsky2024vlms}, logit lens \cite{neo2024towards,huo2024mmneuron}, in-context learning \cite{li2025taco, li2025miv,li2025catp} provide only a coarse-grained analysis of model components or do not fully disentangle the complex interactions between visual and textual token representations. In contrast, our work employs causal tracing with Gaussian noise to precisely quantify the functional roles of MHSA, FFN, and hidden states for both visual and textual tokens across layers, enabling a fine-grained analysis of how LVLMs perceive and process visual object information.
\subsubsection{Mitigating Hallucination in LVLMs}
LVLMs frequently produce content that deviates from visual information, leading to object hallucination. Existing hallucination mitigating strategies can be broadly categorized into three types: \textbf{(1) Training-based approaches} enhance model factuality by pre-training or finetuning with carefully curated datasets \cite{yu2024rlaif,you2023ferret,zhang2025improving} and novel training objectives \cite{lyu2024alleviating}. These methods can be effective but require substantial data and computational resources. \textbf{(2) Contrastive decoding} \cite{leng2024mitigating,huang2024opera,zhong2024investigatingmitigatingmultimodalhallucination} leverages differences between deliberately perturbed decoding paths to promote generations that are more consistent in visual information. However, such methods introduce significant latency at inference time. \textbf{(3) Inference-time interventions} modify internal activations such as attention heads outputs \cite{liu2024paying,li2025cai,ye2025claim} or hidden states \cite{liu2024reducing} to steer the model toward more faithful outputs. 
However, these methods generally lack interpretability of the selection of layers and components.

\begin{figure*}[!t]
\centering
\includegraphics[width=0.89\textwidth]{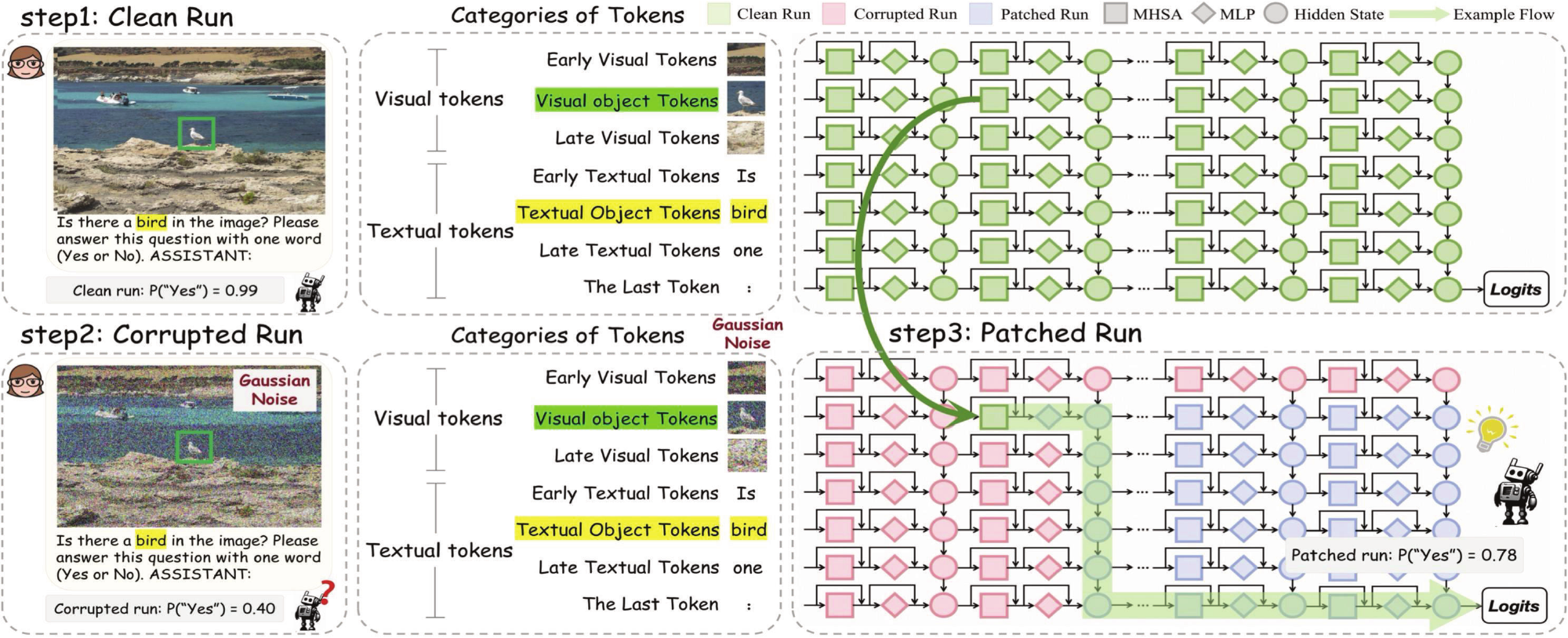} 
\caption{Overview of our proposed \textbf{Fine-grained Cross-modal Causal Tracing} method. Activation patching computes the causal effect of a specific component by running the LVLMs three times: a \textbf{clean run (step1)} with original image, a \textbf{corrupted run (step2)} with image added Gaussian noise, and a \textbf{patched run (step3)} with corrupted input but restoring specific component using the value in the clean run. We use \textbf{Recovery Rate} to quantify the causal effect of each restored component. }
\label{pipeline}
\end{figure*}

\section{Preliminary}
We restrict our scope to LVLMs that are based on auto-regressive Transformer architecture \cite{vaswani2017attention}, as it is adopted by most SOTA LVLMs. The model receives as input a visual input sequence $ \boldsymbol{V} = \{v_1, v_2, \dots, v_m\} $ and a textual input sequence $ \boldsymbol{T} = \{t_1, t_2, \dots, t_n\} $, where $ m $ and $ n $ denote the sequence lengths of the visual and textual input. The textual and visual input sequences are concatenated together and processed through $L$ transformer layers of the language decoder, each consisting of multi-head self-attention (MHSA), feed-forward network (FFN) that is usually a multilayer perception (MLP), and a residual stream is applied between each components. 
The $l$-th layer hidden state $\boldsymbol{h}^{\left(l\right)}$ can be computed from the previous layer:    
\begin{equation}
\boldsymbol{h}^{\left(l\right)} = \boldsymbol{h}^{\left(l-1\right)} + \boldsymbol{a}^{\left(l\right)} + \boldsymbol{m}^{\left(l\right)} ,
\label{eq:attention_output}
\end{equation}
where $\boldsymbol{a}^{\left(l\right)}$ and $\boldsymbol{m}^{\left(l\right)}$ are the output of the MHSA component and the FFN component at layer $l$. Finally, the model predicts the next token in an auto-regressive manner based on the last layer output. 



In this paper, we aim to identify which types of visual and textual tokens, model components (i.e., $\boldsymbol{a}^{(l)}$, $\boldsymbol{m}^{(l)}$, and $\boldsymbol{h}^{(l)}$), and layer ranges play a critical role in the perception and comprehension of visual object information in LVLMs. By uncovering the underlying information flow, we seek to provide practical guidance for mitigating object hallucination and related downstream issues.

\section{Fine-Grained Cross-Modal Causal Tracing}
Causal tracing (also known as activation patching or causal mediation analysis) is a widely used interpretability technique that selectively replaces internal activations to probe the causal contribution of specific model components \cite{meng2022locating}. 
In the context of large language models (LLMs), causal tracing is frequently employed to examine the storage and retrieval mechanisms of factual associations \cite{meng2022locating}, and document-level relevance \cite{liu2025large}. 
In the context of LVLMs, we are the first to propose using controlled Gaussian noise perturbations on input images for causal tracing. By adding controlled Gaussian noise to the entire image and then selectively restoring the activations of specific components, we conduct a fine-grained analysis of the internal mechanisms responsible for visual object perception and comprehension in LVLMs.

Specifically, we select 500 images from the COCO dataset and design object-related questions for the objects present in each image.
Following the analysis methodology of ROME \cite{meng2022locating}, 
we define three types of inference runs:
\begin{itemize}
\item \textbf{Clean Run:} The model is given the original image, and we record the probability $\boldsymbol{P_{\text{clean}}}$ assigned to the token \texttt{yes} in response to binary questions of the form "Is there a XXX in the image? Please answer this question with one word (Yes or No)."
\item \textbf{Corrupted Run:} Gaussian noise is added to the entire image to degrade visual quality, and we record the resulting prediction probability $\boldsymbol{P_{\text{corrupted}}}$.
\item \textbf{Patched Run:} Starting from the corrupted run, we selectively restore specific internal activations (e.g., MHSA output, MLP output, or hidden states) at certain layers and token categories using activations from the clean run. The prediction probability is denoted as $\boldsymbol{P_{\text{patched}}}$.
\end{itemize}

To quantify the causal effect of each restored component, we define the following \textbf{Recovery Rate} ($\boldsymbol{RR}$) metric:
\begin{equation}
    \boldsymbol{RR} =  \frac{\boldsymbol{P_{\text{patched}}}-\boldsymbol{P_{\text{corrupted}}}}{\boldsymbol{P_{\text{clean}}}-\boldsymbol{P_{\text{corrupted}}}}
\label{RR}
\end{equation}

This normalized measure reflects the proportion of clean performance regained through targeted restoration; more profoundly, it serves as a quantitative estimate of the component’s causal effect on visual object perception. A value close to 1 indicates a strong causal effect, whereas a value near 0 suggests minimal influence.


To systematically conduct causal tracing across visual and textual information flow, we define seven categories based on token's position and semantic role in the input sequence:

\begin{figure*}[!t]
\centering
\includegraphics[width=0.98\textwidth]{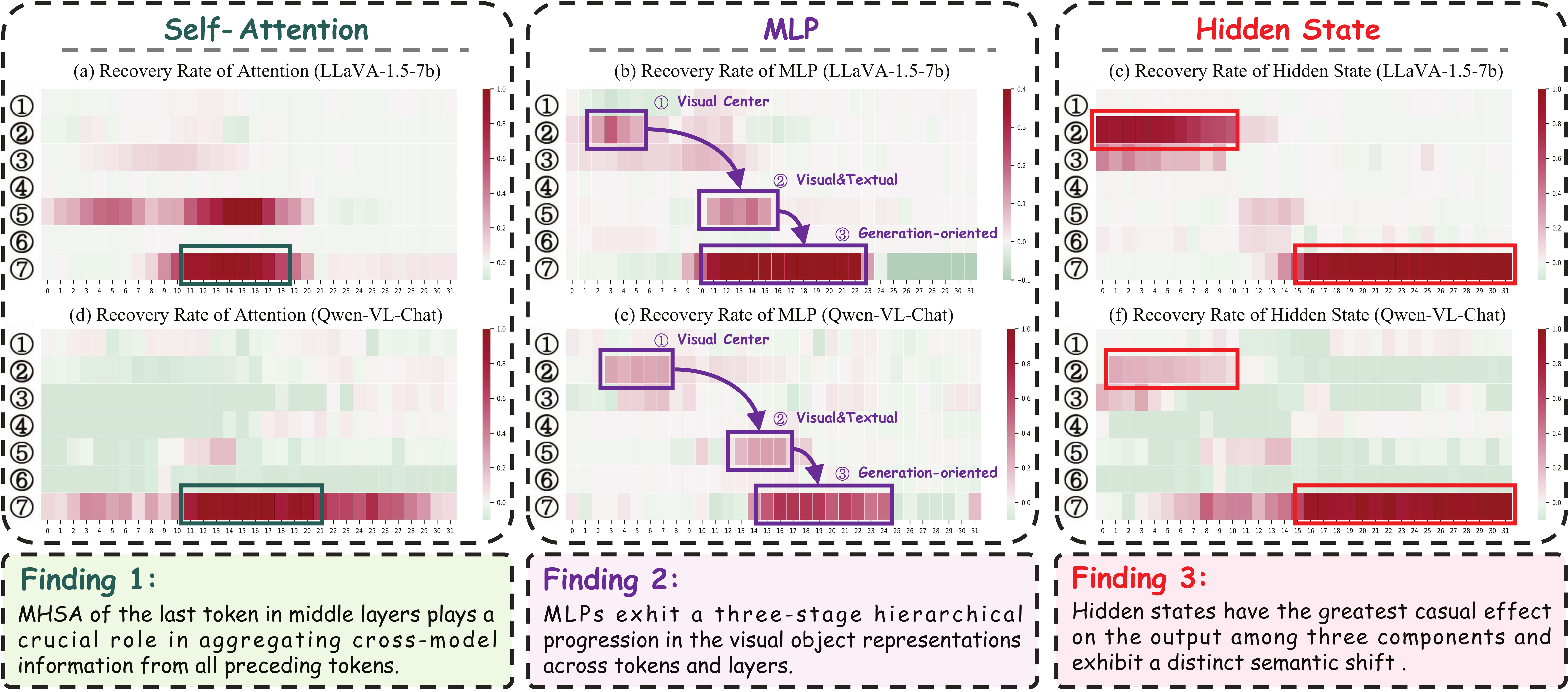} 
\caption{Results and key findings of \textbf{FCCT} framework on LLaVA-1.5-7b and Qwen-VL-Chat. The symbols from \textbf{\circled{1}} to \textbf{\circled{7}} represent the seven token categories defined above: \textbf{\circled{1}\:Early Visual Tokens, \circled{2}\:Object Visual Tokens, \circled{3}\:Late Visual Tokens, \circled{4}\,Early Textual Tokens, \circled{5}\:Textual Object Tokens, \circled{6}\:Late Textual Tokens, and \circled{7}\:The Last Token}.}
\label{findings}
\end{figure*}

\begin{enumerate}
\item[\circlednum{1}] \textbf{Early Visual Tokens} occur before any queried object region, which serve as a control group for comparison.

\item[\circlednum{2}] \textbf{Object Visual Tokens} directly encode visual features corresponding to the queried object, which are central for analyzing the internal mechanisms of visual object perception and comprehension.

\item[\circlednum{3}] \textbf{Late Visual Tokens} occur after any queried object region, which may capture residual visual information.

\item[\circlednum{4}] \textbf{Early Textual Tokens} occur near the visual\&textual sequences boundary, which help analyze how information transitions from visual to language components. 

\item[\circlednum{5}] \textbf{Textual Object Tokens} encode textual features corresponding to the queried object, which help reveal how visual object information interacts with textual reference.


\item[\circlednum{6}] \textbf{Late Textual Tokens} occur in the late part of the textual prompt, which help analyze how visual object information propagates across textual stream not directly related.


\item[\circlednum{7}] \textbf{The Last Token} occurs at the end of input sequence, helping analyze cross-modal information aggregation.
\end{enumerate}


By restoring only one component of one layer for a single token category at one time, we systematically derive fine-grained insights into how LVLMs perceive and comprehend visual object information. This enables us to trace the causal pathways through which visual object information is represented, propagated, and aggregated across different layers and model components.

\subsection{Causal Tracing Results and Key Findings}
In this section, we present and analyze the experimental results and key findings of FCCT conducted on two widely used LVLMs: LLaVA-1.5-7B and Qwen-VL-Chat. As illustrated in Figure~\ref{findings}, we present the \textbf{$\boldsymbol{RR}$s} of 3 model components across 32 layers under 7 token categories, denoted as \circled{\textbf{1}} to \circled{\textbf{7}}, corresponding to the previous definition.

\begin{figure}[!t]
\centering
\includegraphics[width=0.47\textwidth]{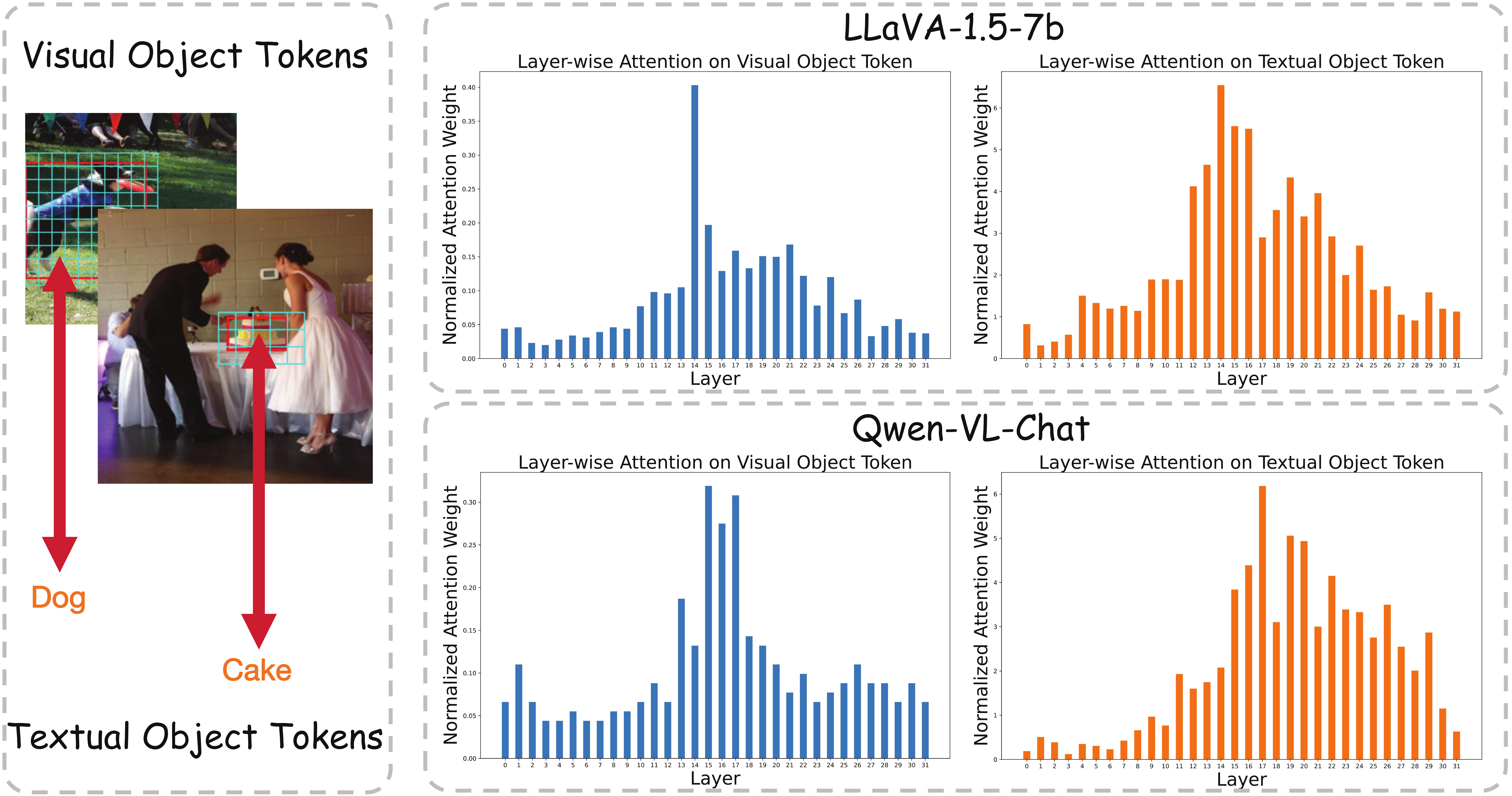} 
\caption{Visualization of normalized attention weights to visual object tokens and corresponding textual object tokens across layers. We report the average result on 3,000 VQAs.}
\label{finding1}
\end{figure}

\subsubsection{Cross-modal Aggregation via the Last Token's MHSAs}
As shown in Figure 2 (a) and (d), the last token's MHSAs in intermediate layers exhibit a strong causal effect, which plays a particularly crucial role in aggregating information from all preceding tokens. 


To further investigate the cross-modal aggregation effect of MHSA, we visualized the last token's layer-wise normalized attention weights for the queried visual object tokens and textual object tokens. As shown in Figure 3, we observe a sharp increase in attention weight around Layer 15. This suggests that around these layers, LVLMs begin to align instruction-guided attention with the most relevant visual and textual cues. We hypothesize that these layers mark a transition point where deep cross-modal information aggregation occurs, enabling the model to bind unimodal representations to high-level cross-modal representations.

\subsubsection{Three-stage Hierarchical Representations via MLPs}

As shown in Figure 2 (b) and (e), MLPs exhibits a three-stage progression in the formation of visual object representations:
In early layers, visual object tokens are encoded into localized, modality-specific embeddings with limited semantic abstraction; In intermediate layers, textual object tokens interact with visual representations, forming increasingly rich cross-modal semantics; In the deeper layers, under the cross-modal aggregation effect of MHSA, the last token's MLPs progressively accumulate a cross-modal and task-relevant representation.

Overall, this hierarchical progression illustrates how MHSA-driven cross-modal interactions and MLPs together transform unimodal and localized visual representations into cross-modal and globally aggregated representations that are essential for visual object perception in LVLMs.

\subsubsection{Hierarchical Semantic Shift of the Hidden States}
As shown in Figure 2 (c) and (f), hidden states exhibit a distinct semantic shift: in shallow layers (Layers 0–10), visual tokens' hidden states are primarily visual-centric, encoding low-level perceptual patterns. In deep layers (Layers 18–31), the last tokens' are cross-modal and highly task-related to the final prediction. Notably, the phase in which the causal effect of the hidden states gradually strengthens aligns with the intermediate layers around Layer 14, precisely where MHSAs and MLPs jointly contribute most significantly to refining cross-model object representations.

This shift reflects the progression from grounding visual object information to aggregation with task-related context for final prediction. It further highlights the hierarchical organization of internal representation inference in LVLMs.


\section{Intermediate Representation Injection}
Motivated by the findings from the proposed FCCT analysis, 
we observe that the last token's MHSA and MLP outputs at intermediate layers are crucial for capturing and aggregating task-related object information from both visual and textual modalities. To take advantage of this observation, 
we propose \textbf{Intermediate Representation Injection (IRI)}, a training-free inference-time technique, which aims to reinforce crucial cross-modal representations to improve visual object information perception and mitigate hallucination.

\subsection{Method}


To reinforce crucial mid-layer cross-modal representations, we selectively inject them into later layers, which are adaptively scaled by their causal effects (Recovery Rates),
thereby amplifying the contributions of the most influential components across layers.

Let $RR^{\text{attn}} = \{RR_k^{\text{attn}}\}_{k=1}^{L}$ and $RR^{\text{mlp}} = \{RR_k^{\text{mlp}}\}_{k=1}^{L}$ denote the Recovery Rates of the MHSA and MLP outputs across the $L$ layers, respectively. 
To ensure that the most critical cross-modal representations are injected into layers with sufficient causal effect, we rank components across layers by their Recovery Rates and independently select:

\begin{itemize}
\item A set of top-$k_1$ MHSA source layers $\mathcal{L}_{\text{src}}^{\text{attn}}$ with the highest $RR^{\text{attn}}$ values and a set of top-$k_2$ target layers $\mathcal{L}_{\text{tgt}}^{\text{attn}}$.
  \item A set of top-$k_1$ MLP source layers $\mathcal{L}_{\text{src}}^{\text{mlp}}$ with the highest $RR^{\text{mlp}}$ values and a set of top-$k_2$ target layers $\mathcal{L}_{\text{tgt}}^{\text{mlp}}$.
\end{itemize}

For each attention source layer $k \in \mathcal{L}_{\text{src}}^{\text{attn}}$, we record the MHSA output of the last token: $\boldsymbol{a}^{(k)} \in \mathbb{R}^d$. For each target layer $l \in \mathcal{L}_{\text{tgt}}^{\text{attn}}$ such that $l > k$, we inject the stored activations into the target MHSA output, and scaling them by $RR$ to modulate their contribution to reflect causal effect:

\begin{equation}
\tilde{\boldsymbol{a}}^{(l)} = \boldsymbol{a}^{(l)} + \lambda_a \cdot \sum_{k \in \mathcal{L}_{\text{src}}^{\text{attn}}} g(k, l)  \cdot RR_k^{\text{attn}} \cdot \boldsymbol{a}^{(k)},
\end{equation}

Similarly, for each $k \in \mathcal{L}_{\text{src}}^{\text{mlp}}$ and $l \in \mathcal{L}_{\text{tgt}}^{\text{mlp}}$ with $l > k$, we inject the recorded MLP outputs as:

\begin{equation}
\tilde{\boldsymbol{m}}^{(l)} = \boldsymbol{m}^{(l)} +  \lambda_m \cdot \sum_{k \in \mathcal{L}_{\text{src}}^{\text{mlp}}} g(k, l) \cdot RR_k^{\text{mlp}} \cdot \boldsymbol{m}^{(k)},
\end{equation}
where $\lambda_a$ and $\lambda_m$ are scaling coefficients, and $g(k, l)$ ensures that the injected information respects causal ordering:

\[
g(k, l) = \begin{cases}
1, & \text{if } l > k \\
0, & \text{otherwise}
\end{cases}
\]

To ensure that the injected activation maintains the same norm as the original, we apply the following normalization:

\begin{equation}
\tilde{\boldsymbol{a}}^{(l)} = \tilde{\boldsymbol{a}}^{(l)} \cdot \frac{\lVert \boldsymbol{a}^{(l)} \rVert_2}{\lVert \tilde{\boldsymbol{a}}^{(l)} \rVert_2}, \tilde{\boldsymbol{m}}^{(l)} = \tilde{\boldsymbol{m}}^{(l)} \cdot \frac{\lVert \boldsymbol{m}^{(l)} \rVert_2}{\lVert \tilde{\boldsymbol{m}}^{(l)} \rVert_2},
\end{equation}
where $|| \cdot ||$ represents the $\ell_2$ norms (Euclidean norms) of the activation vectors.

This intervention ensures that only critical cross-modal intermediate activations are injected into subsequent layers with sufficient causal effect throughout the information flow, enhancing the LVLM's trustworthiness to visual object information and thereby mitigating hallucination.

\subsection{Experimental Setup}
\subsubsection{Models.} We adopt the widely used LLaVA-1.5-7b \cite{liu2024improved}, Qwen-VL-Chat \cite{bai2023qwen}, LLaVA-NeXT \cite{liu2024llavanext}, Qwen2-VL-7B\cite{wang2024qwen2} and InternVL2-8B \cite{chen2024far} as baseline LVLMs.
\subsubsection{Evaluation.}
We comprehensively evaluate the methods for both discriminative and generative tasks to measure the effectiveness and robustness of hallucination mitigation.
\begin{itemize}
    \item \textbf{POPE} \cite{li2023evaluating} employs a binary question-answering format, inquiring LVLMs to answer if a special object exists in the given image. Following previous works, we adopt Accuracy and F1 score as the metrics.
    \item \textbf{MME} \cite{Fu2023MMEAC} serves as a comprehensive tool for assessing the capabilities of LVLMs across both 10 perception tasks and 4 cognition tasks. Consequently, task scores are reported as the evaluation metric.
    \item \textbf{CHAIR} \cite{rohrbach2018object} is a widely used metric for assessing object hallucination in responses of LVLMs. 
    The CHAIR metric comprises two important indicators, denoted as ${C}_S$ and ${C}_I$, with the following calculation formulas:
    \begin{equation}
\small
   C_S = \frac{|\{\textrm{Hallucinated}\ \textrm{objects}\}|}{|\{\textrm{All}\ \textrm{mentioned}\ \textrm{objects}\}|} \notag
\end{equation}
\begin{equation}
\small
   C_I = \frac{|\{\textrm{Sentences w/ hallucinated objects}\}|}{|\{\textrm{All sentences}\}|} \notag
\end{equation}

\noindent
    \item \textbf{MMHal-Bench} \cite{sun2023aligning} comprises 96 meticulously designed questions, which evaluates response-level hallucination rate (VH.\%) and informativeness (Score). It asks \textbf{GPT-4} to compare model outputs with human responses and object labels for evaluation.
    \item \textbf{MHumanEval} \cite{yu2024rlaif} is designed to evaluate hallucination performance by \textbf{human annotators}. The benchmark contains 146 samples collected from Object HalBench and MMHal-Bench. Given model responses, we ask three human annotators to label the hallucinated segments and compute the mean response-level hallucination rate (Hu.\%) as the evaluation metric.
\end{itemize}

\subsubsection{Baselines.}
We compared our proposed \textbf{IRI} method with the following SOTA training-free methods:
\textbf{VCD} \cite{leng2024mitigating} contrasts model logits derived from original and distorted visual input to reduce the over-reliance on statistical bias and unimodal priors. \textbf{OPERA} \cite{huang2024opera} introduces a penalty term on the model logits during the beam-search decoding to mitigate the over-trust issue. \textbf{PAI} \cite{liu2024paying} intervenes on attention heads by leveraging their original direction and optimizes the output distribution during decoding to mitigate language bias. \textbf{VTI} \cite{liu2024reducing} mitigates hallucination by steering layer-wise hidden states during inference to enhance visual feature stability.

\begin{table*}[!ht] 
\small
\renewcommand{\arraystretch}{0.60} 
\setlength{\tabcolsep}{4.5pt} 
\centering
\begin{tabular}{lccccccccccccccc}
\toprule
\multirow{2}{*}{\textbf{Method}}  & \multicolumn{5}{c}{\textbf{LLaVA-1.5-7b}} & \multicolumn{5}{c}{\textbf{Qwen-VL-Chat}} & \multicolumn{5}{c}{\textbf{LLaVA-NeXT}} \\
\cmidrule(lr){2-6} \cmidrule(l){7-11} \cmidrule(l){12-16}
& Exist. & Count & Pos. & Color &Total& Exist. & Count & Pos. & Color & Total & Exist. & Count & Pos. & Color & Total \\
\midrule 
\rowcolor{gray!20}Regular &175.7& 124.7  & 114.0 & 151.0&565.4 & 170.0  & 135.0 & 123.3 & 170.0& 598.3 &180.0 &105.0 &150.0 & 151.7 & 586.7  \\
VCD&180.3&131.7  & 125.0 & 155.0 & 592.0   & \underline{180.0}&133.3&131.7&175.0& 620.0 & 185.0 &125.0 & 133.3 & 168.3 &611.6  \\
OPERA&165.0& 116.0 & 133.3 & 149.0 & 563.3  &  \underline{180.0} &140.0&\textbf{138.3}&175.0&633.3 & 183.8 & 121.3 & \textbf{155.0} &162.1& 622.2 \\
PAI &\underline{190.0} & \textbf{148.3} &126.7 & 160.0 & 625.0 & 175.0 & 141.6 & 132.5 & 177.5 & 626.6 & 185.0 & \underline{128.3} & 148.3 & 170.8 & 632.4 \\
VTI & 185.0 & \underline{140.0} &\underline{135.0} &\underline{165.7} & \underline{625.7} & \underline{180.0} & \underline{142.5} & 133.0 & \underline{178.0} & \underline{633.5} & \underline{186.7} & 126.7 & \underline{150.0} & \underline{172.5} & \underline{635.9} \\
\textbf{IRI(ours)}&\textbf{195.0}&\underline{140.0}  &\textbf{140.0}  &\textbf{168.3}  & \textbf{648.3}  & \textbf{185.0} &\textbf{145.0}&\underline{135.0}&\textbf{180.0}&\textbf{645.0} & \textbf{190.0} & \textbf{135.0}& \textbf{155.0 }&\textbf{177.5}& \textbf{657.5} \\
\bottomrule
\end{tabular}
\caption{Results on MME hallucination subset. The best performances are bolded and the second-best are underlined.  }
\label{tab:mme_result}
\end{table*}

\begin{table}[!ht]
\small
\renewcommand{\arraystretch}{0.9}
\centering
\setlength{\tabcolsep}{0.5pt}

\begin{tabular}{clcccccc}
\toprule
\multirow{2}{*}{\textbf{Setting}} & \multirow{2}{*}{\textbf{Method}} & \multicolumn{2}{c}{\textbf{LLaVA-1.5-7b}} & \multicolumn{2}{c}{\textbf{Qwen-VL-Chat}} & \multicolumn{2}{c}{\textbf{LLaVA-NeXT}}\\
\cmidrule(lr){3-4} \cmidrule(l){5-6} \cmidrule(l){7-8}
& & Acc & F1 & Acc & F1 &Acc & F1\\
\midrule
\multirow{6}{*}{Ran.} 
&\cellcolor{gray!20} Regular & \cellcolor{gray!20}83.29 & \cellcolor{gray!20}81.33 & \cellcolor{gray!20}84.63 & \cellcolor{gray!20}82.61 & \cellcolor{gray!20}84.78 & \cellcolor{gray!20}86.43 \\
& VCD & 87.73 & 87.16 & 86.93 & 85.46 & 88.76 & 89.57 \\
& OPERA & 89.20 & 88.81 & 85.71 & 84.64 & 90.27 & 89.71 \\
& PAI & 86.33 & 84.56 & 85.38 & 85.54 & 88.40 & 87.16 \\
& VTI & 89.50 & 88.89 & 86.73 & 85.59 & 89.23 & 88.68 \\
& \cellcolor{green!15} \textbf{IRI(ours)} & \cellcolor{green!15}\textbf{89.76} & \cellcolor{green!15}\textbf{89.32} & \cellcolor{green!15}\textbf{87.38} & \cellcolor{green!15}\textbf{87.42} & \cellcolor{green!15}\textbf{90.68} & \cellcolor{green!15}\textbf{90.21} \\
\cmidrule(lr){2-8}
\multirow{6}{*}{Pop.}
& \cellcolor{gray!20} Regular & \cellcolor{gray!20}81.88 & \cellcolor{gray!20}80.06 & \cellcolor{gray!20}83.63 & \cellcolor{gray!20}81.53 & \cellcolor{gray!20}83.23 & \cellcolor{gray!20}84.77 \\
& VCD & 85.38 & 85.06 & 85.17 & 83.68 & 87.01 & 87.70 \\
& OPERA & 86.64 & 86.62 & 84.82 & 83.99 & 87.16 & 87.68 \\
& PAI & 85.33 & 83.62 & 84.20 & 83.10 & 86.65 & 86.99 \\
& VTI & 87.36 & 86.69 & 85.67 & 84.48 & 87.33 & 87.16 \\
& \cellcolor{green!15}\textbf{IRI(ours)} & \cellcolor{green!15}\textbf{87.67} & \cellcolor{green!15}\textbf{87.07} & \cellcolor{green!15}\textbf{86.24} & \cellcolor{green!15}\textbf{86.89} & \cellcolor{green!15}\textbf{88.25} & \cellcolor{green!15}\textbf{88.04}\\
\cmidrule(lr){2-8}
\multirow{6}{*}{Adv.}
& \cellcolor{gray!20} Regular & \cellcolor{gray!20}78.96 & \cellcolor{gray!20}77.57 & \cellcolor{gray!20}81.03 & \cellcolor{gray!20}79.30 & \cellcolor{gray!20}81.19 & \cellcolor{gray!20}82.50 \\
& VCD & 80.88 & 81.33 & 83.10 & 82.04 & 84.80 & 85.23 \\
& OPERA & 81.24 & 81.38 & 82.67 & 79.89 & 85.20 & 85.54 \\
& PAI & 83.17 & 81.67 & 82.19 & 82.06 & 84.32 & 83.68 \\
& VTI & 82.57 & 82.11 & 83.13 & 82.16 & 85.35 & 84.52 \\
& \cellcolor{green!15}\textbf{IRI(ours)} & \cellcolor{green!15}\textbf{85.17} & \cellcolor{green!15}\textbf{84.18} & \cellcolor{green!15}\textbf{84.83} & \cellcolor{green!15}\textbf{84.52} & \cellcolor{green!15}\textbf{85.67} & \cellcolor{green!15}\textbf{86.26}\\
\bottomrule
\end{tabular}
\caption{Results on POPE tasks. We evaluate the accuracy and F1 Score of various widely used LVLMs. 
}
\label{tab:main_result}
\end{table}

\begin{table}[!ht] 
\small
\centering
\renewcommand{\arraystretch}{0.90} 
\setlength{\tabcolsep}{2pt} 
\begin{tabular}{lcccccccc}
\toprule
\multirow{2}{*}{ \textbf{Method}} & \multicolumn{4}{c}{\textbf{LLaVA-1.5-7b}} & \multicolumn{4}{c}{\textbf{Qwen-VL-Chat}} \\
\cmidrule(lr){2-5}
\cmidrule(lr){6-9}
& $C_S$ \textbf{$\downarrow$} & $C_I$ \textbf{$\downarrow$} & Recall$\uparrow$& Len& $C_S$ \textbf{$\downarrow$} &  $C_I$ \textbf{$\downarrow$} &Recall$\uparrow$ & Len \\
\midrule
\rowcolor{gray!20}Regular & 52.8 & 15.9 &77.3 &93.4& 2.8 & 3.0&31.0&5.3 \\
VCD & 51.0 & 14.9 &77.2& 101.9& 1.4 & 1.2&30.8&4.0  \\
OPERA & 45.6 & 13.1& \textbf{78.5}&95.3 & 1.7 & 1.3&31.9 &4.4\\
PAI & 38.3 & 12.4 & 76.9 &94.4 &  1.3 & 1.2&\underline{32.2}&4.2 \\
VTI & \underline{36.9} & \underline{12.1} &76.8&93.8 &\underline{1.1} &\underline{1.1}&31.4 &4.2\\
\textbf{IRI(ours)} & \textbf{34.6} & \textbf{11.5} &\underline{78.2}&95.8&  \textbf{1.0} & \textbf{0.9} &\textbf{32.6} &4.4 \\
\bottomrule
\end{tabular}
\caption{Results on CHAIR. $Max\ new\ tokens$ is 512. Lower $C_S$ and $C_i$ along with higher recall and length indicate better hallucination mitigating performance.
}
\label{Chair-result}
\end{table}

\subsubsection{Implementation Details.}
In our experiments, we uniformly set $k_1 = 3$ and $k_2 = 10$. For LLaVA-1.5-7B, we use $\lambda_a = 0.26$ and $\lambda_m = 0.16$; for Qwen-VL-Chat, $\lambda_a = 0.20$ and $\lambda_m = 0.10$; and for LLaVA-NeXT, $\lambda_a = 0.15$ and $\lambda_m = 0.08$.
Both causal tracing analysis and evaluation experiments of our proposed IRI method are performed on 8 $\times$ NVIDIA A100 SXM 80GB GPUs.

\subsection{Main Results}

Based on the experimental results presented in Tables 1-5, we can draw the following key conclusions:

\begin{table}[!ht] 
\small
\centering
\renewcommand{\arraystretch}{0.90} 
\setlength{\tabcolsep}{2pt} 
\begin{tabular}{lcccccc}
\toprule
\multirow{2}{*}{ \textbf{Method}} & \multicolumn{3}{c}{\textbf{LLaVA-1.5-7b}} & \multicolumn{3}{c}{\textbf{Qwen-VL-Chat}} \\
\cmidrule(lr){2-4}
\cmidrule(lr){5-7}
& Score\textbf{$\uparrow$} & VH.\%\textbf{$\downarrow$}& Hu.\%\textbf{$\downarrow$}& Score\textbf{$\uparrow$} & VH.\%\textbf{$\downarrow$}&Hu.\%\textbf{$\downarrow$} \\
\midrule
\rowcolor{gray!20}Regular & 1.86 & 63.5& 67.1 & 2.93 & 41.1&61.0 \\
VCD & 2.12 & 54.2 &66.7&  2.77 & 39.2 & 61.5\\
OPERA & 2.15 & 54.2&63.0 & 2.94 & 38.4 &58.2\\
PAI & 2.27 & 53.2 &62.5&  2.87 & 39.5 & 56.7 \\
VTI & 2.43 & 52.2 &63.4& 2.99 & 38.4 &57.4\\

\rowcolor{green!15}\textbf{IRI(ours)} & \textbf{2.53} & \textbf{50.2} & \textbf{62.0} &\textbf{3.13} & \textbf{37.5} & \textbf{56.2} \\
\bottomrule
\end{tabular}
\caption{Results on MMHal-Bench and MHumanEval. We use \textbf{GPT-4} and \textbf{human annotators} as evaluation references.
}
\label{mmhbench-result}
\end{table}

\begin{table}[!t] 
\small
\centering
\renewcommand{\arraystretch}{1} 
\setlength{\tabcolsep}{2pt} 
\begin{tabular}{rccccccc}
\toprule
\multirow{2}{*}{ \textbf{Model\ \ \ \ \ \ }} & \multicolumn{2}{c}{\textbf{POPE}} & \multicolumn{2}{c}{\textbf{MME}} & \multicolumn{2}{c}{\textbf{CHAIR}} \\
\cmidrule(lr){2-3}
\cmidrule(lr){4-5}
\cmidrule(lr){6-7}
&ACC \textbf{$\uparrow$} & F1 \textbf{$\uparrow$}
& Cog.\textbf{$\uparrow$} & Hall.\%\textbf{$\uparrow$} & $C_S$ \textbf{$\downarrow$} &  $C_I$ \textbf{$\downarrow$} \\
\midrule
\rowcolor{gray!20}Qwen2-VL-7B & 88.49 & 87.85 & 556.4 & 630.0& 24.8&7.2 \\
\rowcolor{green!15}\textbf{+ IRI} & \textbf{89.04} & \textbf{88.44} &  \textbf{563.4} & \textbf{663.3}& \textbf{14.2}&\textbf{6.5} \\
\midrule
\rowcolor{gray!20}InternVL2-8B & 86.67 & 85.72 & 566.4 & 663.0 & 37.2&9.4\\
\rowcolor{green!15}\textbf{+ IRI} & \textbf{87.69} & \textbf{86.90} &  \textbf{569.3} & \textbf{688.7} & \textbf{30.7} &\textbf{8.6}  \\
\bottomrule
\end{tabular}
\caption{Results on more advanced models. Cog. and Hall. denote the cognitive and hallucination subset of MME.
}
\label{advance-result}
\end{table}

\textbf{(1) Robust and SOTA Performance:} Our proposed IRI method demonstrates robust, SOTA hallucination mitigation performance across both discriminative and generative tasks. Specifically, on the POPE benchmark, IRI achieves an average improvement of +4.89\% in Accuracy and +5.20\% in F1 Score. For the MME hallucination subset, IRI brings an average absolute gain of +65.7 points in the Total score.
On the CHAIR benchmark, IRI reduces the average hallucination metrics ($C_S$ and $C_I$) by 6.43 points.
Finally, on MMHal-Bench, IRI improves the average Score by +0.44 while lowering the average VH Rate by 8.45\%.
These results demonstrate the effectiveness and robustness of our approach in mitigating hallucinations.

\textbf{(2) Model-agnostic and generalizable:} IRI is not dependent on specific model architectures and can be readily deployed across various LVLMs. We successfully implemented IRI on more advanced models, such as Qwen2-VL-7B and InternVL2-8B, where it continued to provide steady and significant performance enhancements.

\textbf{(3) Preserving foundational capabilities:} IRI effectively mitigates hallucination without sacrificing LVLM's other foundational capabilities. Specifically, it leads to improved scores on MME cognitive tasks and more informative responses, as indicated by higher scores on MMHal-Bench.

\begin{table}[!t] 
\small
\centering
\renewcommand{\arraystretch}{0.3} 
\setlength{\tabcolsep}{2pt} 
\begin{tabular}{lccccccc}
\toprule
\multirow{2}{*}{ \textbf{Setting\ \ \ \ \ \ }} & \multicolumn{4}{c}{\textbf{Hyperparameters}} & \multicolumn{2}{c}{\textbf{POPE}} \\
\cmidrule(lr){2-5}
\cmidrule(lr){6-7}
&$\lambda_a$ & $\lambda_m$
& $k_1$ & $k_2$ & ACC \textbf{$\uparrow$} & F1 \textbf{$\uparrow$} \\
\midrule
\rowcolor{gray!20}\textbf{LLaVA-1.5-7b} & - & - & - & - &78.96 &77.57 \\
\midrule
\multicolumn{7}{c}{{\textit{Ablation of Component}}}\\
\midrule
+ IRI w/o MLP&0.22 & - & \textbf{3} & \textbf{10}& 84.93& 84.02\\
+ IRI w/o MHSA& - & 0.04 & \textbf{3} & \textbf{10}& 85.07 & 84.03 \\
+ IRI w/ Hidden States& 0.18 & 0.04 & \textbf{3} & \textbf{10}& 84.50 & 83.67 \\
\midrule
\multicolumn{7}{c}{{\textit{Ablation of Layer Range}}}\\
\midrule
+ IRI w/ Fisrt 10 Layers&0.22 & 0.14 & \textbf{3} & \textbf{10}& 78.46 & 77.23 \\
+ IRI w/ Last 10 Layers&0.20 & 0.14 & \textbf{3} & \textbf{10}& 80.42& 79.87 \\
\midrule
\multicolumn{7}{c}{{\textit{Ablation of Layer Nums.}}}\\
\midrule
+ IRI w/ Less Source Layers & 0.26 &0.16 & 1 &\textbf{10} &82.92&83.13 \\
+ IRI w/ More Source Layers & 0.24 &0.14 & 5 &\textbf{10} &84.76&83.94 \\
+ IRI w/ Less Target Layers & 0.22 &0.12 & \textbf{3} &5 &83.32&82.71 \\
+ IRI w/ More Target Layers & 0.26 &0.16 & \textbf{3} &15 &84.49&83.56 \\
\midrule
\multicolumn{7}{c}{{\textit{Ablation of $RR$s}}}\\
\midrule
+ IRI w/o $RR$s &0.26 & 0.14 & \textbf{3} & \textbf{10}& 84.42& 83.92 \\
\midrule
\multicolumn{7}{c}{{\textit{Ablation of Normalization}}}\\
\midrule
+ IRI w/o Norm.&0.24 & 0.14 & \textbf{3} & \textbf{10}& 85.07& 83.98 \\
\midrule
\rowcolor{green!15}\textbf{+ IRI} & 0.26 & 0.16 &  \textbf{3} & \textbf{10} & \textbf{85.17}&\textbf{84.18} \\
\bottomrule
\end{tabular}
\caption{Result of ablation study on MS-COCO POPE. For each experiment, the parameter $\lambda_{a}$ and $\lambda_{m}$ is individually optimized to ensure fair comparison. }
\label{ablation-result}
\end{table}

\subsection{Ablation Study}
As shown in Table \ref{ablation-result}, to validate the effectiveness of each component within the proposed IRI method and the key findings of FCCT framework, we conducted a comprehensive and systematic set of ablation experiments. We specifically focus on addressing the following five questions:

\noindent\textbf{\textit{(1) Why is it necessary to intervene in both MHSA and MLP, but not directly in hidden states?}}
Experimental results show that removing either MHSA or MLP results in a slight performance decrease. Combined aggregation of two modules yields greater improvements, which demonstrates that both MHSAs and MLPs play a critical role in enriching the high-level representations in the middle layers. Furthermore, we also apply interventions to the hidden states based on IRI and observe a performance drop. 
We believe that hidden states are the cumulative result of MHSA, MLP, and the states from the previous layer. Directly intervening in these highly integrated and functionally specialized hidden states can disrupt the hierarchically constructed flow of semantic information. Consequently, this approach is less effective than precisely enhancing the individual key components: the MHSA, which is responsible for information aggregation, and the MLP, which handles representation processing.

\noindent\textbf{\textit{(2) Do the intermediate layers really play a crucial role?}} Specifically, when IRI's source and target layers are both limited in the first ten or the last ten layers, accuracy drops to 78.46\% and 80.42\%, respectively. These results align well with our FCCT findings: intermediate layers carry the strongest causal effect for perceiving and aggregating crucial visual\&textual object information, 
whereas shallow layers lack sufficient crucial visual object perception and deep layers already focus on final output generation. 

\noindent\textbf{\textit{(3) How does the number of source and target layers affect performance?}} 
We find that the number of intervention layers affects the effectiveness of IRI to some extent.
It is necessary to select a sufficient number of intermediate layers with strong causal effects for visual information perception and inject them into a sufficient number of target layers to make IRI effective. Notably, even when injecting from too many source layers or into too many target layers, IRI's performance does not drop significantly compared to its peak, which demonstrates strong robustness. Nevertheless, selecting too many source layers may introduce noise, while including too many target layers may inject mid-level representations into task-relevant representation for final prediction, preventing optimal performance.

\noindent\textbf{\textit{(4) Why is Recovery Rate necessary for more precise injection?}}  
The Recovery Rate adaptively controls how much each layer’s information is amplified in line with its estimated causal effect. By re-weighting restored activations, it highlights components across layers that contribute more strongly to visual object perception.

\noindent\textbf{\textit{(5) Why is normalization necessary for more stable injection?}}  
The normalization strategy ensures that the scale of the vector remains consistent before and after injection, preventing undesired magnitude shifts that may distort downstream representations. This mechanism stabilizes the effect of representation injection, thereby enhancing the robustness of the IRI method to distributional shifts.


\subsection{Inference Latency}

As shown in Table~\ref{lantency}, IRI achieves the best hallucination mitigating performance while preserves the inference speed.

\begin{table}[htbp] 
\small
\centering
\renewcommand{\arraystretch}{0.3} 
\setlength{\tabcolsep}{8pt} 

\begin{tabular}{lrrc}
\toprule
\textbf{Method} & \textbf{TTFT(ms)} &\textbf{ TPOT(ms)} & \textbf{Acc(\%)}\\
\midrule
\rowcolor{gray!20} LLaVA-1.5-7b & 99.8 {{1.0}\(\times\)} & 36.0 {1.0\(\times\)} & 78.96 \\
\midrule
+ VCD & 160.1 {1.6\(\times\)} & 96.8 {2.7\(\times\)} & 80.88 \\
+ OPERA & 109.8 {1.1\(\times\)} & 69.5 {1.9\(\times\)} & 81.24\\
+ PAI & 156.3 {1.6\(\times\)} & 93.6 {2.6\(\times\)} & 83.17 \\
\midrule
\rowcolor{green!15}+ \textbf{IRI(ours)} & 102.2 {1.0\(\times\)} & 36.5 {1.0\(\times\)}  & 85.17 \\
\bottomrule
\end{tabular}
\caption{Inference latency (Time to First Token, Time Per Output Token) and the accuracy on adversarial POPE.
}
\label{lantency}
\end{table}

\section{Conclusion}

In this paper, we introduce the \textbf{Fine-grained Cross-modal Causal Tracing (FCCT)} framework and the \textbf{Intermediate Representation Injection (IRI)} technique to improve the interpretability and performance of large vision-language models (LVLMs). Our FCCT framework provides a comprehensive, fine-grained causal analysis of the internal components of LVLMs, uncovering key insights into the cross-modal aggregation and hierarchical representation formation, particularly through MHSA and MLP mechanisms. Building on these insights, IRI proves to be a robust and training-free inference-time method, significantly mitigating object hallucinations across various LVLM architectures while maintaining inference speed and foundational model capabilities. Experimental results not only demonstrate the superiority of IRI, 
but also validate 
the findings of FCCT.

\section{Acknowledgement}
Xiaocheng Feng and Xiachong Feng are the co-corresponding authors of this work. We thank the anonymous reviewers for their insightful comments. This work was supported by the National Natural Science Foundation of China (NSFC) (grant 62522603, 62276078, U22B2059), the Key R\&D Program of Heilongjiang via grant 2022ZX01A32, and the Fundamental Research Funds for the Central Universities ( XNJKKGYDJ2024013 ).

\bibliography{aaai2026}

\end{document}